\documentclass{article}

\usepackage{arxiv}

\usepackage[utf8]{inputenc} % allow utf-8 input
\usepackage[T1]{fontenc}    % use 8-bit T1 fonts
\usepackage{hyperref}       % hyperlinks
\usepackage{url}            % simple URL typesetting
\usepackage{booktabs}       % professional-quality tables
\usepackage{amsfonts}       % blackboard math symbols
\usepackage{nicefrac}       % compact symbols for 1/2, etc.
\usepackage{microtype}      % microtypography
\usepackage{lipsum}		% Can be removed after putting your text content
\usepackage{float}
\usepackage{graphicx}
\usepackage{natbib}
\usepackage{doi}

\title{Shedding some light on light up with Artificial Intelligence}

\author{ 
    Libo Sun \\
	Samuel Ginn College of Engineering \\
    Auburn University \\
	Auburn, AL USA \\
	\texttt{lzs0101@auburn.edu} \\
	\And
	James Browning \\
	Samuel Ginn College of Engineering \\
    Auburn University \\
	Auburn, AL USA \\
	\texttt{jlb0181@auburn.edu} \\
	\AND
	\href{https://orcid.org/0000-0001-8688-780X}{\includegraphics[scale=0.06]{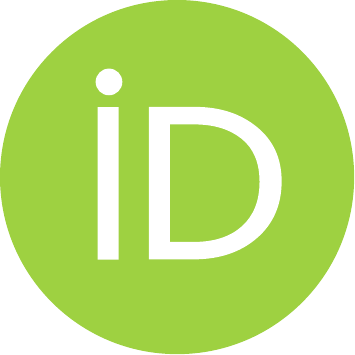}\hspace{1mm}Roberto Perera} \\
	Department of Aerospace Engineering \\
    Auburn University \\
	Auburn, AL USA \\
	\texttt{rzp0063@auburn.edu} \\
}

\renewcommand{\shorttitle}{Shedding some light on light up with Artificial Intelligence}

\hypersetup{
pdftitle={Shedding some light on light up with Artificial Intelligence},
pdfsubject={cs.artificial-intelligence},
pdfauthor={Libo Sun, James Browning, Roberto Perera},
pdfkeywords={artificial intelligence, hill climbing search, simulated annealing, supervised machine learning, deep neural networks, light up puzzle, akari},
}

\begin{document}
\maketitle

\begin{abstract}
	The Light-Up puzzle, also known as the AKARI puzzle, has never been solved using modern artificial intelligence (AI) methods. Currently, the most widely used computational technique to autonomously develop solutions involve evolution theory algorithms. This project is an effort to apply new AI techniques for solving the Light-up puzzle faster and more computationally efficient. The algorithms explored for producing optimal solutions include hill climbing, simulated annealing, feed-forward neural network (FNN), and convolutional neural network (CNN). Two algorithms were developed for hill climbing and simulated annealing using 2 actions (add and remove light bulb) versus 3 actions (add, remove, or move light-bulb to a different cell). Both hill climbing and simulated annealing algorithms showed a higher accuracy for the case of 3 actions.
    The simulated annealing showed to significantly outperform hill climbing, FNN, CNN, and an evolutionary theory algorithm achieving $100 \%$ accuracy in 30 unique board configurations. Lastly, while FNN and CNN algorithms showed low accuracies, computational times were significantly faster compared to the remaining algorithms. The GitHub repository for this project can be found at  https://github.com/rperera12/AKARI-LightUp-GameSolver-with-DeepNeuralNetworks-and-HillClimb-or-SimulatedAnnealing.
\end{abstract}

% keywords can be removed

\section{Introduction}
As far as we can tell, the Light Up puzzle has never been solved using modern artificial intelligence methods (or, at the very least, it has never been documented). The methods currently used to find solutions to this puzzle involve evolution theory. Therefore, we want to solve the Light Up puzzle using various new AI techniques.

We believe vanilla hill climbing and simulated annealing can prove to be simple and efficient approaches to solving this puzzle. When compared to evolution theory, these two methods are much easier to understand and implement. For example, by creating an evaluation function which rewards the moves that reduce search space and punish moves that cause restrictions, a much simpler approach can be formulated.

Additionally, these methods allow us to easily integrate constraint functions that follow through the initialization step as well as finding the solution. For instance, a constraint which avoids breaking the rules of the game of non-intersecting bulbs and restrictions of adjacent bulbs to the black cells. Lastly, these methods also allow us to avoid redundant moves such as moving/removing a bulb which might reduce the reward. Therefore, by exploring these methods, we can offer a new, attractive, and easier path to solving the Light Up puzzle.

In addition to the vanilla hill climbing and simulated annealing, the next AI method proposed to be explored is the use of neural networks. We believe neural networks could be extremely useful for solving this puzzle in the future. For example, they would grant users the ease of not relying on a step-by-step search for solutions of the puzzle by outputting an entirely solved puzzle directly from the initial configuration. Additionally, neural networks have proved to be extremely fast and computationally efficient, requiring a low amount of processing memory. Therefore, by researching into new neural network approaches to solve the Light Up puzzle, we can overcome the current challenges of evolution theory algorithms.

Therefore, this project will focus on solving the Light Up puzzle starting from informed search algorithms like hill climb and simulated annealing, and then moving to neural network algorithms like feedforward neural networks and convolutional neural networks.

%\begin{center}
\begin{figure}
  \includegraphics[width=1.0\linewidth]{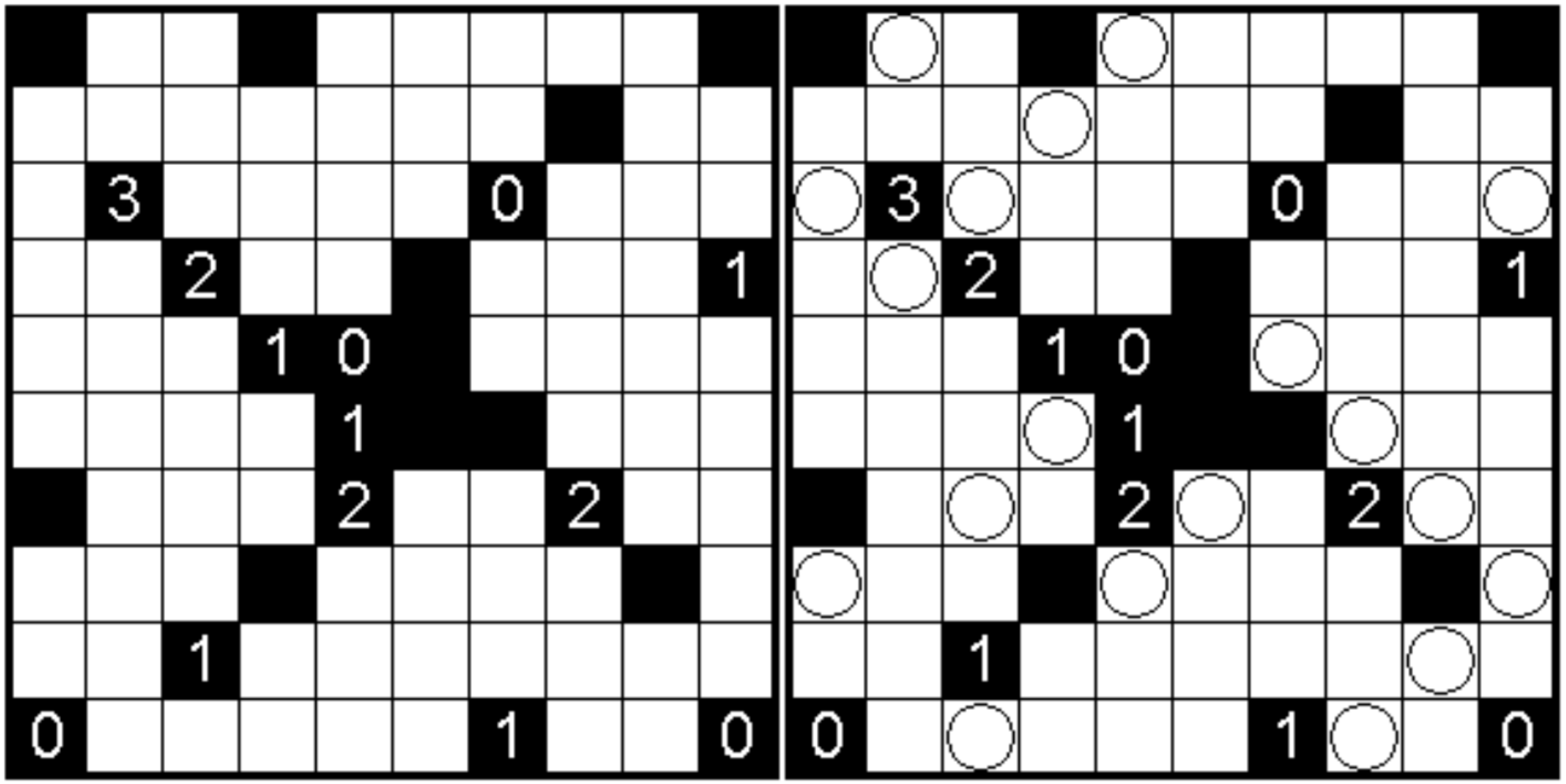}
  \centering
  \caption{Demonstration of an initial Light Up board configuration (left), along with its initialized configuration (right).}
\end{figure}
%\end{center}

\section{Methods}
\subsection{Game/Puzzle rules}
The game works as follows:

\begin{enumerate}
    \item Each puzzle begins with a rectangular grid of black and white cells.
    \item Solving the puzzle involves placing light bulbs into white squares to illuminate the entire board without any overlapping rays.
    \item Black squares can’t be filled, and black squares with numbers inside them must have that number of adjacent bulbs.
    \item Players must place light bulbs in white cells, in an arrangement such that… \\
        (a) Every white square is illuminated (light bulbs send rays of light up, right, down, and left, unless blocked by a black square). \\
        (b) No two light bulbs shine on each other.
    \item Players must pay attention to the numbers inside some black squares, which impose a limit on how many light bulbs can be placed in adjacent white squares.
\end{enumerate}

\subsection{Approach}
\begin{itemize}
    \item \textbf{Initial states:} The boards will be generated at random using user-defined inputs, such as board dimensions and number of black cells.
    \item \textbf{Approach:} First, light bulbs will be placed at random initial positions, based on the given limit of adjacent bulbs for each black square. The board may contain some initial violations (such as two bulbs shining on each other).
    \item \textbf{Output:} The system will attempt to produce an optimal solution, based on the performance measure, using several different search algorithms. Allowed moves will include placing a bulb on an empty cell and removing a bulb from a filled cell.
    \item \textbf{Algorithms:} First, we will implement informed search algorithms — including various forms of hill climbing, as well as simulated annealing — to determine what moves to make. Second, after the simulated annealing method has been implemented successfully, a large data set of randomly generated initial states, as well as the moves needed to solve them, will be saved by running the simulated annealing for approximately one week. Ideally, this will generate enough training data to make training a deep neural network possible, with the goal of solving the puzzle given any initial state. Our definition of performance will include how many white squares a move lights up (increasing performance), as well as how many violations it might cause (decreasing performance).
    \item \textbf{Baselines:} Baselines will include various methods of hill climbing, simulated annealing, and deep neural networks. To compare their performance, we plan to analyze factors such as the total average time it takes for each method to solve a series of puzzles, the average number of moves made, and the average performance cost. It will be extremely important to explore various tuning parameters for each algorithm, such as temperature and probability for simulated annealing, and number of hidden layers, number of neurons per hidden layer, learning rate, and activation functions (e.g. cross entropy, sigmoid) for deep neural networks. Additionally, other game-specific parameters must be tuned, such as the number of neighbors which are generated for selection, conditions of search termination, and choice of hill climbing variant.
    \item \textbf{Completeness:} According to research by Brandon McPhail \citep{first}, the light up puzzle is NP-complete, though it is not discussed nearly as much in artificial intelligence cycles as problems like eight queens—making it a perfect problem to solve for a project like this.
\end{itemize}

\subsection{Algorithm's puzzle formalization}

The first step of developing this project in earnest was to formalize the Light Up puzzle into a format that could be understood by an AI — to this end, we chose to represent boards as matrices of single digits, with each representing a possible state for the space. They are as follows:

\begin{itemize}
\item[] CELL\_BLACK\_ZERO = 0
\item[] CELL\_BLACK\_ONE = 1
\item[] CELL\_BLACK\_TWO = 2
\item[] CELL\_BLACK\_THREE = 3
\item[] CELL\_BLACK\_FOUR = 4
\item[] CELL\_BLACK\_FIVE = 5
\item[] CELL\_EMPTY = 6 (This value indicates an empty white cell.)
\item[] CELL\_BULB = 7 (This value indicates a white cell with a bulb inside it.)
\item[] CELL\_LIGHT = 8 (This value indicates that a cell is lit up by a bulb.)
\item[] CELL\_BULB\_ZERO = 9 (This value indicates that a bulb cannot be placed adjacent to this cell.)
\end{itemize}

With these definitions, any conceivable Light Up puzzle can be translated into a format readable by our program as shown in Figure 2.

%\begin{center}
\begin{figure}
    \centering
    \includegraphics[width=0.4\textwidth]{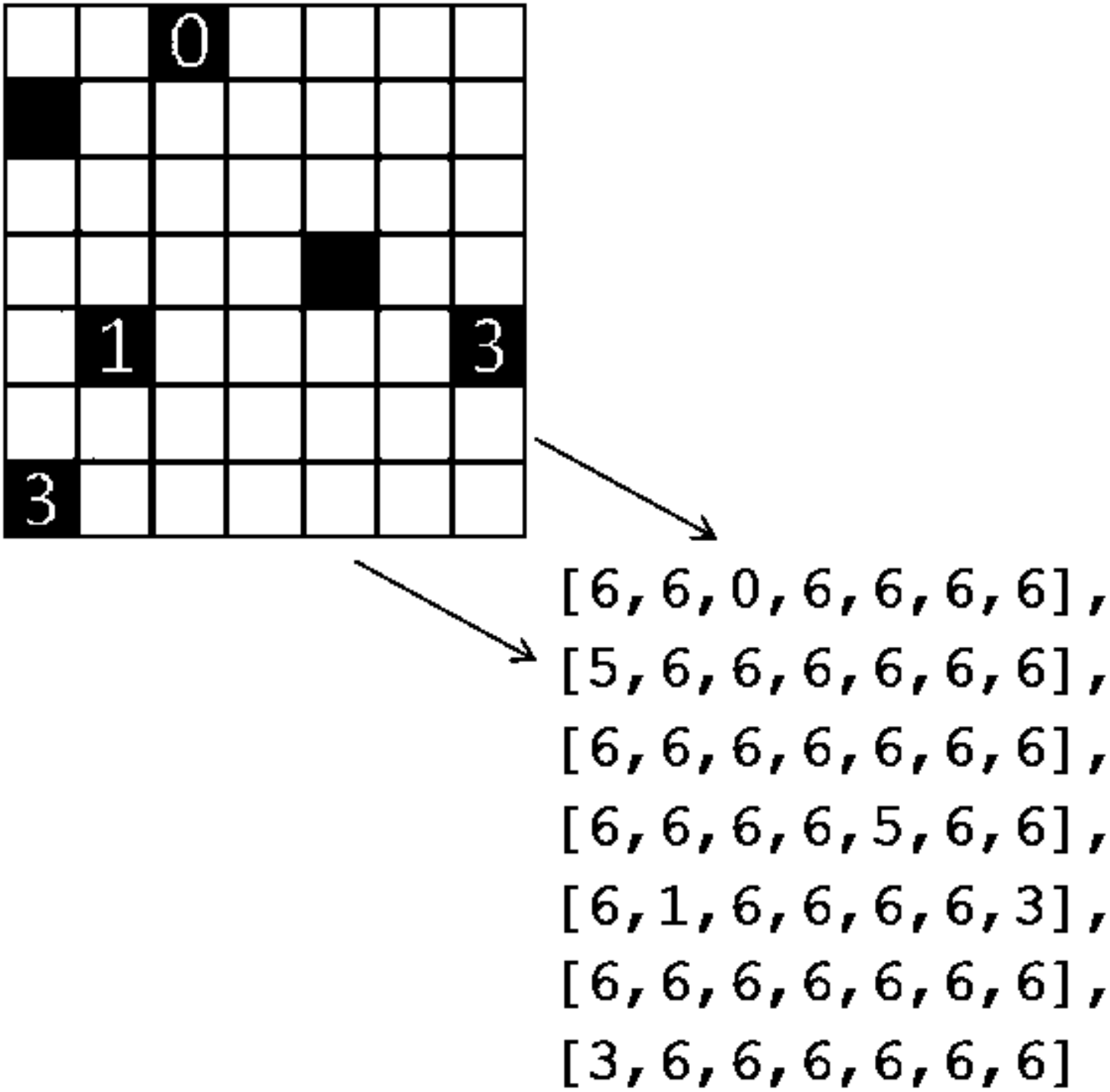}
    \centering
    \caption{Demonstration of an initial Light-Up board with its corresponding numerical array.}
    \centering
\end{figure}
%\end{center}

\subsection{Hill Climbing Algorithms and Simulated Annealing methodology}

The methodology of the developed search algorithms, hill climbing and simulated annealing is shown in the flowchart of Figure 3.
The step-by-step process will explained as follows:

\begin{figure}
    \centering
    \includegraphics[width=\textwidth]{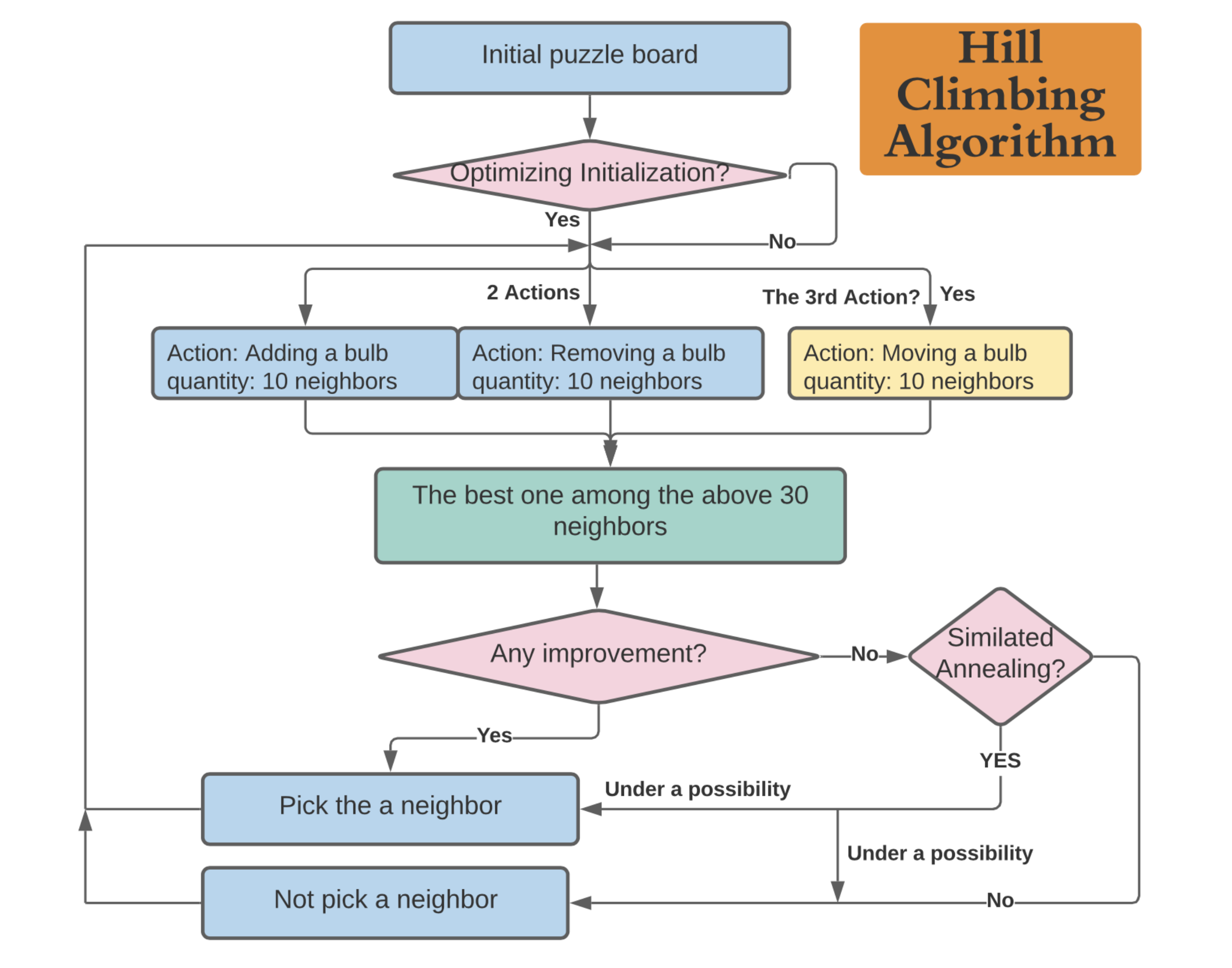}
    \centering
    \caption{ Flowchart of the developed Hill-Climb algorithm and Simulated Annealing framework.}
    \centering
\end{figure}

\begin{enumerate}
    \item Actions:
        \begin{enumerate}
            \item Adding a bulb: Insert a bulb into an empty cell,then calculate the corresponding fitness.
            \item Removing a bulb: Try to remove a bulb from the current board and generate a corresponding fitness
            \item Moving a bulb: Moving a bulb from the current position to another empty cell and generate a fitness then.
        \end{enumerate}
    \item {Fitness:} The percentage of the number of cells that are lit up, and is penal by the number of conflicts: two bulbs shining each other or the number of black adjacency is not met.
    \item Random normal initialization: Blindly add bulbs to meet the black cell adjacency.
    \item Optimizing initialization. Some cells may only have unique way for placing bulbs in, thus the searching space can be reduced.
    \item Hill climbing explanation:\\
    By generating a set of neighbors, pick a best one among those neighbor. If the best neighbor is not better than the current status, the simulated annealing will have some possibility to pick the no good neighbor \citep{third, fourth}, otherwise the hill climbing will not pick a neighbor.
    \item Evaluations for data analysis: By using 30 runs of 100000 evaluations, we have got the minimum requirements for t-test.
    \item {Baselines:}
        \begin{enumerate}
            \item Vanilla hill climbing: plain hill climbing
            \item plus optimized initialization: Use an optimized to narrow the search space.
            \item plus simulated annealing: allow the search to pick a worse neighbor under a possibility
            \item 3 actions VS 2 actions
        \end{enumerate}
    \item Search Space estimated: 
    
    The search space estimated for the board shown in Figure \ref{fig:search_space} is: \\ 
    2 to the power of (100-25) $\rightarrow$ $2^{75}$ = 
377 7893186295 7161709568.

    \begin{figure}
        \centering
        \includegraphics[width=0.3\textwidth]{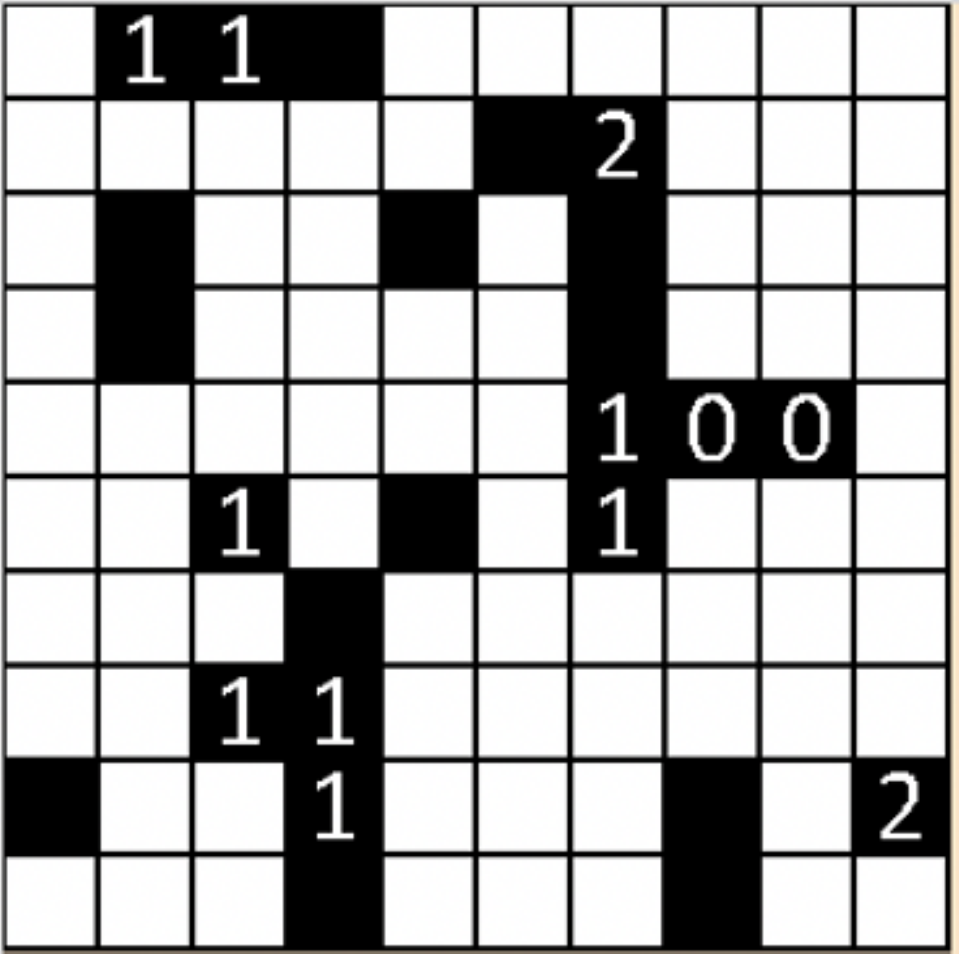}
        \caption{Random board used to compute search space.}
        \label{fig:search_space}
        \centering
    \end{figure}

\end{enumerate}

\section{Results for Hill Climbing and Simulated Annealing}
A guide to executing the hill climbing code and examining the outputs can be found in the appendix. 

\subsection{Results of performance measure}
    Figure 5 shows the architecture used to evaluate the performance of the search algorithms.
    Additional details for each algorithm can be found in the Supplementary Section.
    Furthermore, using the procedure shown in Figure 5 the performance of each algorithm was computed and compared to a conventional evolutionary algorithm which we call Optimized Initialization.
    The resultant performance shown in Figure 6 was developed by evaluating the percentage of lit up cells for a total of 30 unique boards. 
    A key observation to make is that the simulated annealing algorithm outperformed the hill climbing and the optimized initialization algorithms achieving $100 \%$ accuracy for all cases.
    
    \begin{figure}
        \centering
        \includegraphics[width=0.9\textwidth]{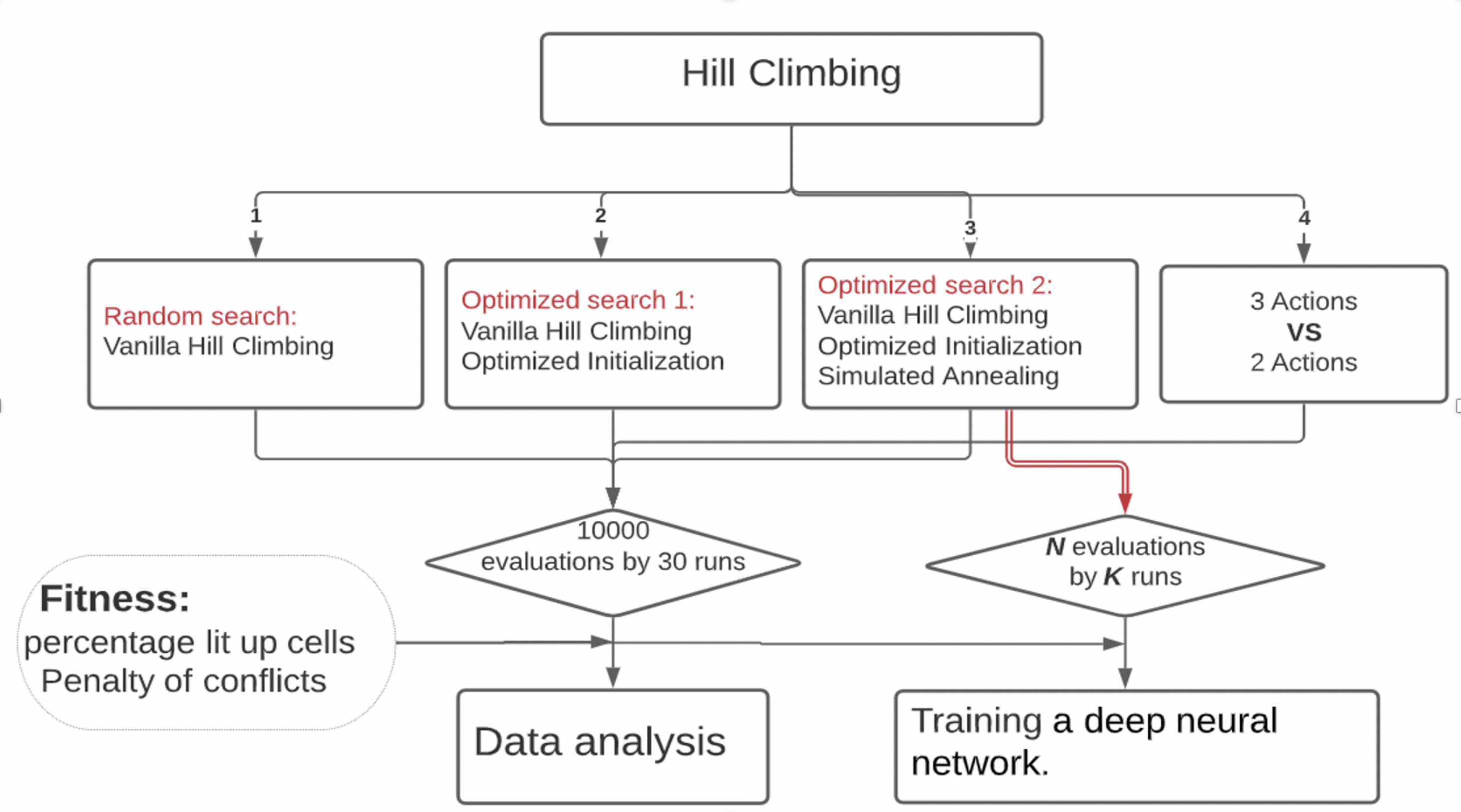}
        \caption{Flowchart of the Hill Climb and Simulated Annealing fitness evaluation procedure.}
        \centering
    \end{figure}

    \begin{figure}
        \centering
        \includegraphics[width=0.9\textwidth]{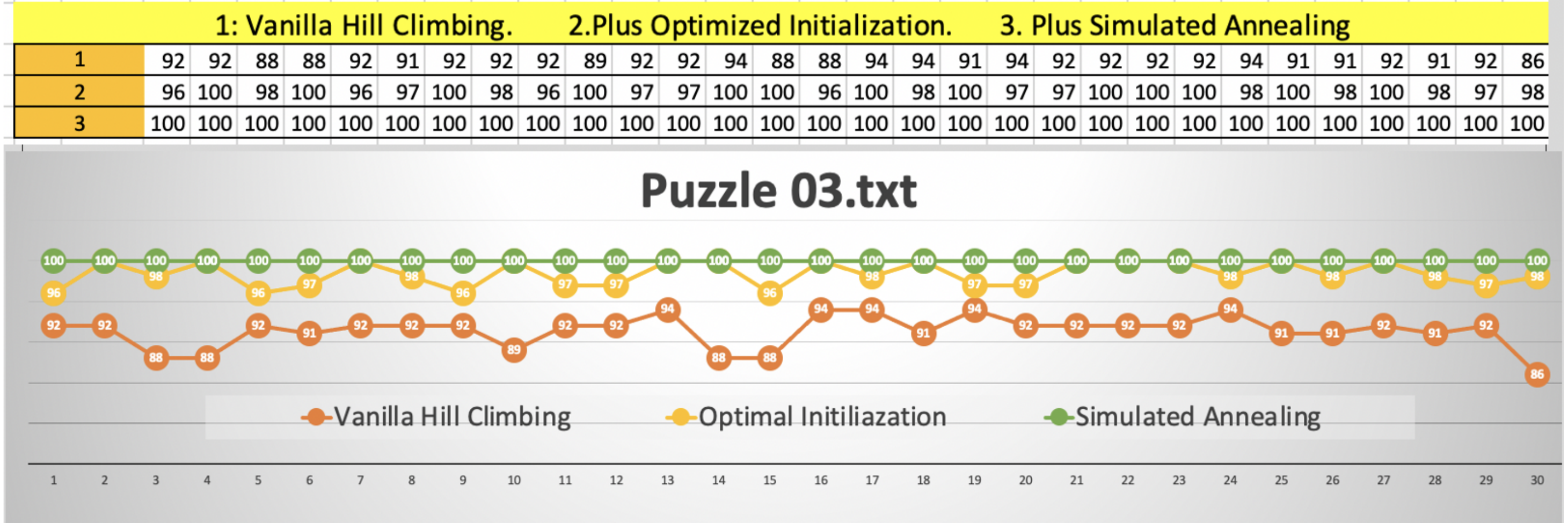}
        \caption{Performance measure results for the Search Algorithms with 30 unique cases.}
        \centering
    \end{figure}
    
\subsection{Varying parameters of T-test and P-test}
    The next test performed for the hill climbing and simulated annealing was to test the algorithm's accuracy when varying the temperature parameter (T-test) and probability parameter (P-test). 
    As shown below in Figures 7-8, a T-Test was performed to validate the results from Figure 6 \citep{t-test}. As expected, Figures 7-8 show to validate the same performance relationship obtained from Figure 6. Note that the P-value in the tables denotes a measure of the probability that an observed difference could have occurred just by random chance. The lower the P-value, the greater the statistical significance of the observed difference, hence, the better the performance of the corresponding algorithm. Additionally, it is important to emphasize that in order to perform the T-test, a minimum of 30 values in a single group needed to be generated, as shown in Figure 6. 
    Therefore, because both P-values are much less than 0.05, we can statistically conclude that the simulated annealing has the best performance, while the optimized initialization has the second best and the vanilla hill climbing is the worst.
    \begin{figure}
        \centering
        \includegraphics[width=0.9\textwidth]{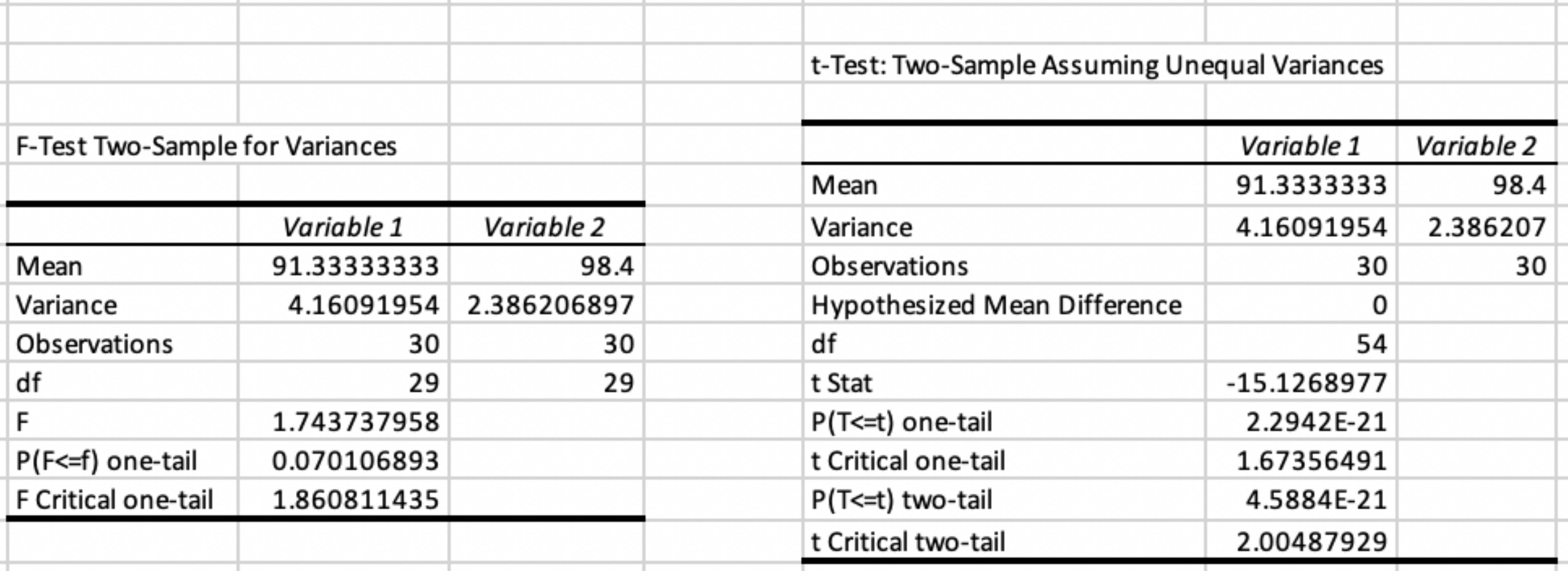}
        \caption{F-test and T-test performance measures for hill climbing by varying and non-varying Variances.}
        \centering
    \end{figure}
    
    \begin{figure}
        \centering
        \includegraphics[width=0.9\textwidth]{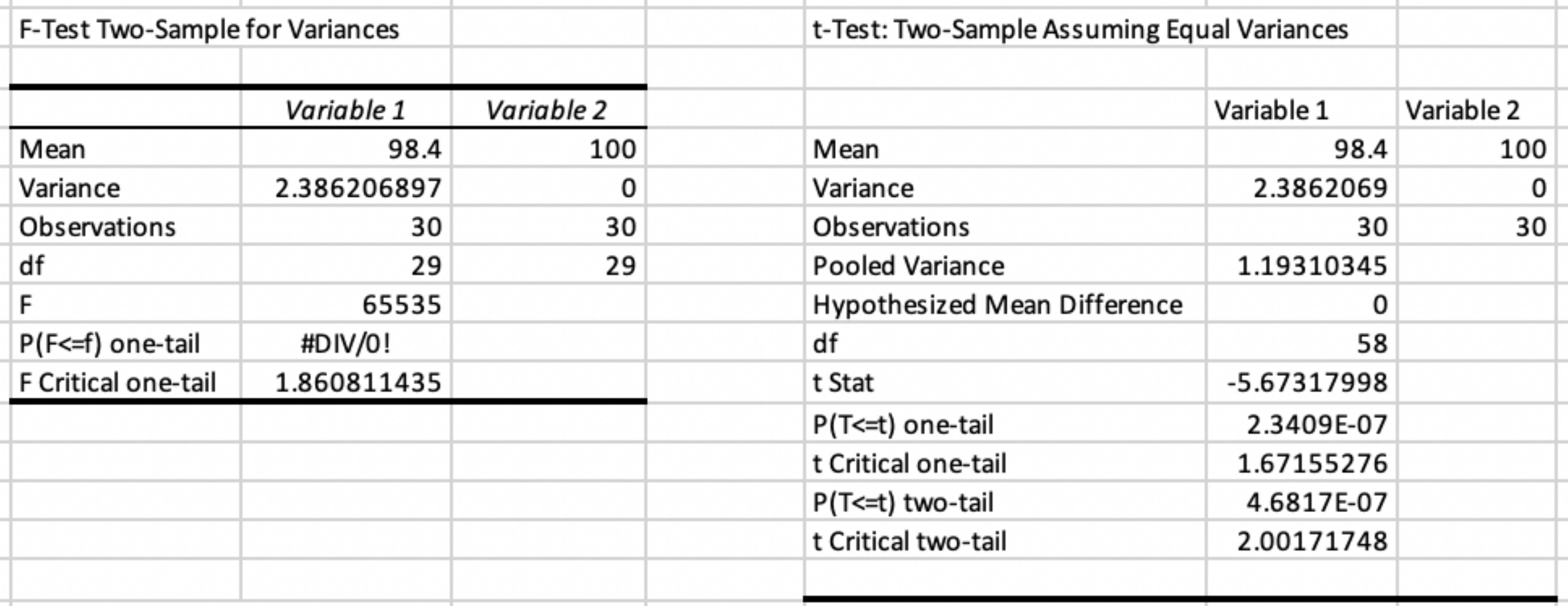}
        \caption{F-test and T-test performance measures for simulated annealing by varying and non-varying Variances.}
        \centering
    \end{figure}
    
\subsection{2 actions versus 3 actions}   

    \begin{figure}
        \centering
        \includegraphics[width=0.3\textwidth]{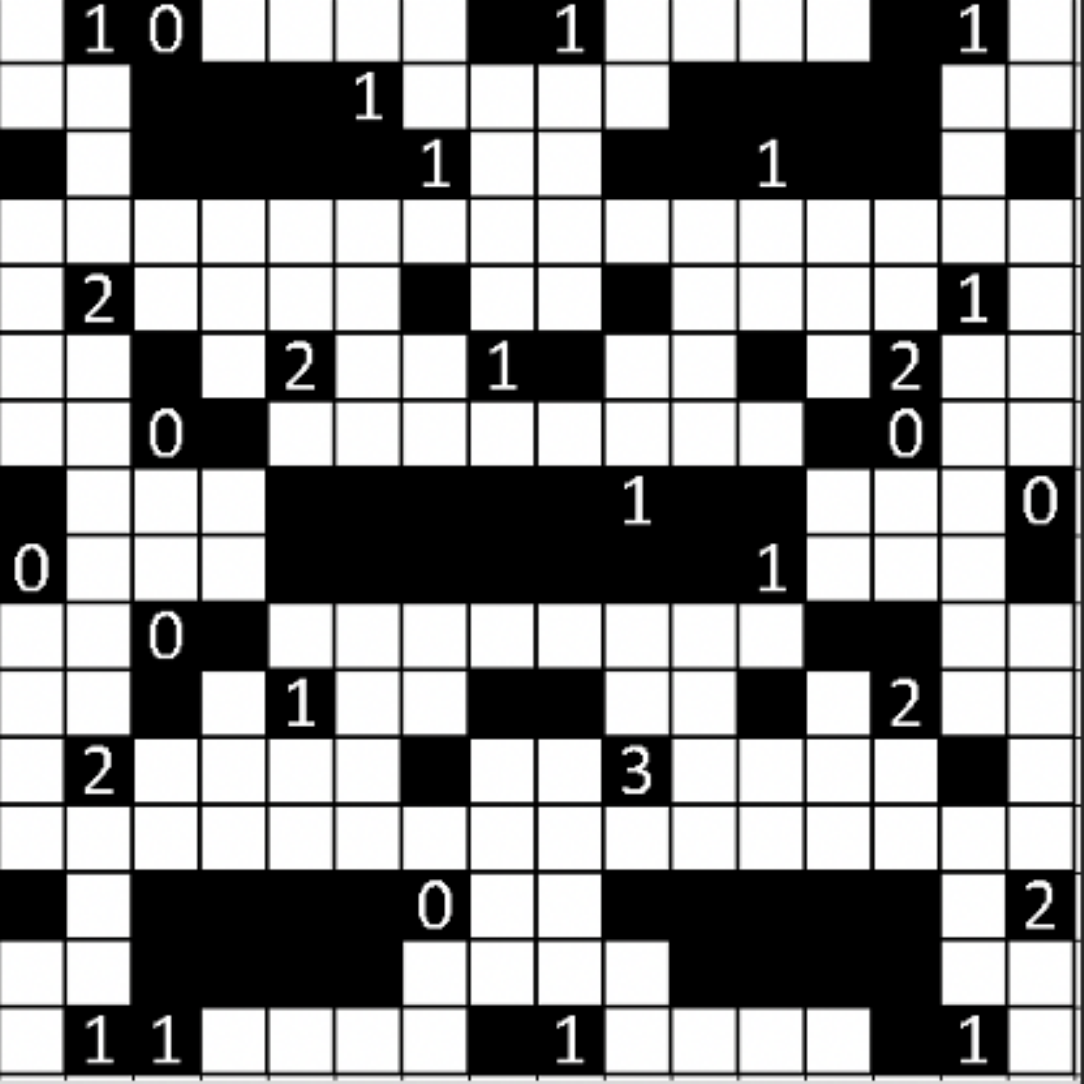}
        \caption{16x16 board used for the performance analysis of 3 actions vs. 2 action.}
        \centering
    \end{figure}

    The final evaluation of the hill climbing and simulated annealing algorithms was to compare the performance of each method by varying from 2 actions (remove or add a light bulb) versus 3 actions (remove, add, or move a light bulb to a different cell). 
    For this analysis, a random higher-complexity board consisting of 16 by 16 cells was generated as shown in Figure 9.
    As expected, using three actions showed a better performance versus only using 2 actions. It can be deducted that by allowing one more action by moving a bulb to a different cell, the performance can be improved for certain cases. As shown below in Figure 10, the running results depict this outcome for all 30 unique boards.
    Additionally, in Figure 11 a T-Test was performed for the 30 cases generated for 3 actions in order to validate the findings depicted in Figure 10.
    From the results reported for the Hill Climb and Simulated Annealing analysis we can see from Figure 11 that since the P-Value is 4.11 E-08, hence, much smaller than 0.05 obtained in the T-Test shown in Figures 7 and 8, we can statistically say that the 3 actions case resulted in higher accuracies compared to the 2 actions case.
    This observation is also reinforced from figure 10 where the accuracies for the case of 3 actions were higher in most cases.
    
    \begin{figure}
        \centering
        \includegraphics[width=\textwidth]{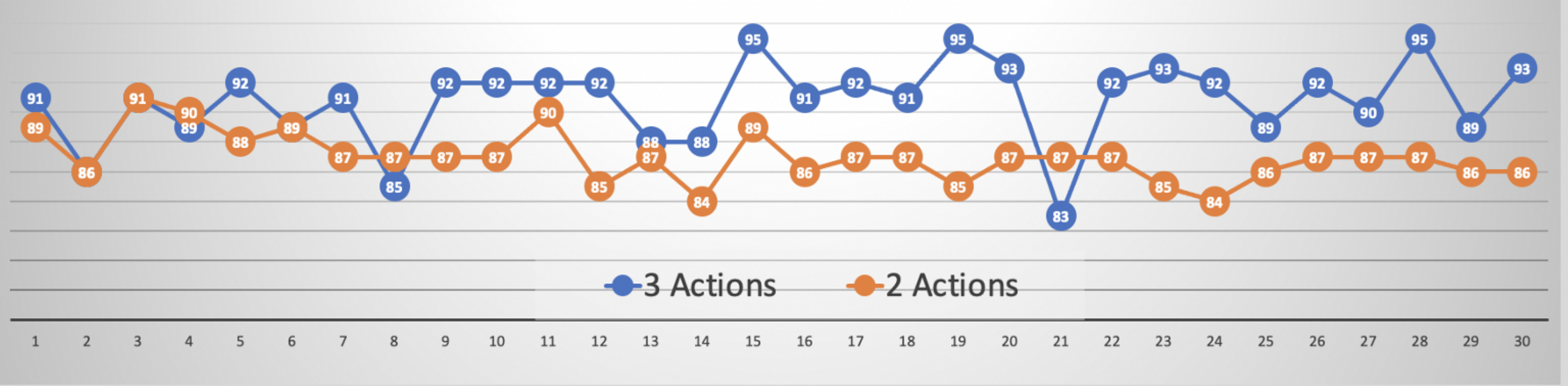}
        \caption{Results for 3 actions versus 2 actions using 30 unique cases.}
        \centering
    \end{figure}

    \begin{figure}
        \centering
        \includegraphics[width=0.9\textwidth]{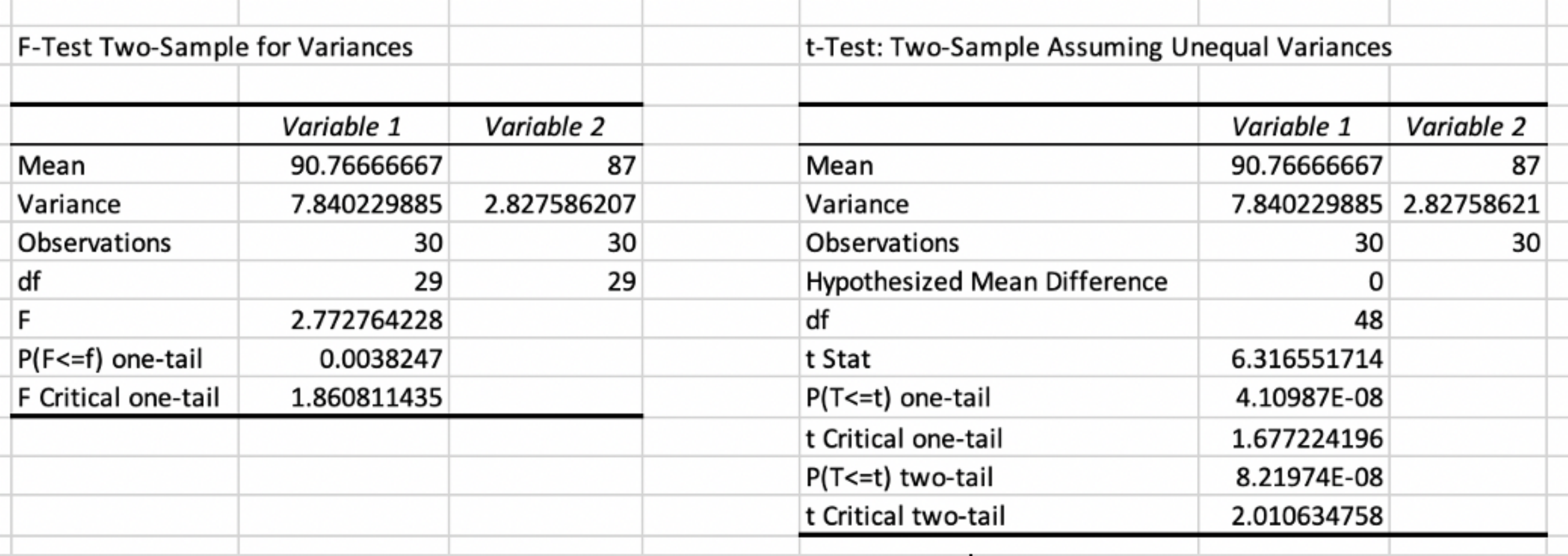}
        \caption{T-Test performance measures by varying and non-varying Variances of the cases for 3 actions versus 2 actions.}
        \centering
    \end{figure}
 
\section{Neural Networks}
\subsection{\normalsize{Methodology for Deep Neural Networks}}

The Light-up Puzzle (also known as AKARI Puzzle) offers a unique challenge to Artificial Intelligence (AI), specifically, when implementing neural networks to generate a solution. The main challenges of the puzzle involve not only the large search space which depends on the initial board dimensions and available black cells, but on the variety of different configurations that result in a solution for the board. 
Additionally, another important challenge to mention is that the Light-Up puzzle does not offer the feasibility to be solved using simple constraint functions integrated within a sequential neural network such as RNN. For example, unlike the "Sudoku Puzzle" which has a single optimal solution configuration for each board, and where the player can solve the puzzle moving one cell at a time allowing the use of simple constraint functions integrated to a sequential network such as RNNs \citep{sudoku}, the Light-Up puzzle does not involve such features. Instead, the solution to the Light-Up puzzle requires to first solve the initialized board by placing adjacent bulbs to each black cell without a specific order, and second, to place light-bulbs which illuminate the remaining white cells while meeting the game rules. These requirements not only restrict the use of a sequential network since the game cannot be solved one cell at a time in ascending order, but also causes a near impossible challenge when finding a global optima even with the use of intricate constraint functions. 
Lastly, the computational search for solutions to the Light-Up puzzle has never been attempted with neural networks. Typically, due to the large search space coupled with the problems of locating the global optima, the methods attempted are those seen in Evolution Theory \citep{evolution}. Thus, this only increased the difficulty of the project by requiring the development of data sets for the Vanilla Hill-Climb initializations, Vanilla Hill-Climb full solutions, and Simulated Annealing solutions. As a result of these unique challenges, this puzzle offered our group with more fun and attractive problem for exploring the use of FNNs and CNNs. 

The following baselines proposed to solve the "Light Up Puzzle" are Deep Neural Networks (DNN) and Convolutional Neural Networks (CNN). Three groups of data sets of randomly initialized puzzles with dimensions 7x7 along with their respective solutions were generated. The first data set consisted of 5000 random initial boards with their respective Vanilla Hill-Climb initialization (i.e. black cell with No. 3 can contain up to 3 adjacent light-bulbs) requiring approximately 24 hours to gather. Once the full Vanilla Hill-Climb algorithm was developed, a second data set of 5000 randomly initialized boards with their solutions was generated in approximately 3-4 days. Additionally, a data set of 10,000 randomly initialized boards was also developed for the simulated annealing algorithm requiring approximately 5-6 days. Lastly, all data sets developed were split using 80$\%$ for the training set, and 20$\%$ for the validation set.
Three neural network approaches were then tested on the data sets including one Feedforward Neural Network (FNN), and two different architectures of CNNs. A key detail to point out is that the preferred deep learning platform used in all of the following methods was "PyTorch" due to it's feasibility of researching multiple architectures and variations. The results showed that the FNN was not able to capture the correct relationship between the initial boards and the solution to the puzzle by misplacing light bulbs and crosses, and changing the structure of the initial state. Lastly, while the CNNs were able to capture a stronger relationship towards the solution by recognizing that all white cells needed to become illuminated without altering the initial state configuration, it failed to capture rules such as light-bulbs being the source of illumination, two light bulbs not being able to intercept, and black cells being constrained to a specific number of adjacent light bulbs.    
Ultimately, while CNNs are limited to develop a generalization of the puzzle's rules as well as over-fitting, it is a promising approach to explore in the future using more advanced neural networks techniques. A promising strategy for future exploration can be to include constrain functions as well as a sequential algorithm, often seen in Recurrent Neural Networks (RNN) \citep{RNN}.

\subsection{\normalsize{Feedforward Neural Networks (FNN)}}

The first implementation of neural networks to solve the light-up puzzle was a FNN following a similar architecture as shown in Figure 12. to solve the initialized boards pertaining to the Vanilla Hill-Climb initialization algorithm. The initialization algorithm was performed to 5000 randomly generated boards to be used as a training set where each random board represented the input to the neural network, and the solved initialized board represented the labeled set to be predicted. In other words, for each random initial state, light bulbs were placed adjacent to the black cells according to their constraint or black cell value. The generation of this dataset took approximately 24 hours to gather and to upload to Auburn University Hopper workstation. 

\begin{figure}
  \includegraphics[width=0.4\linewidth]{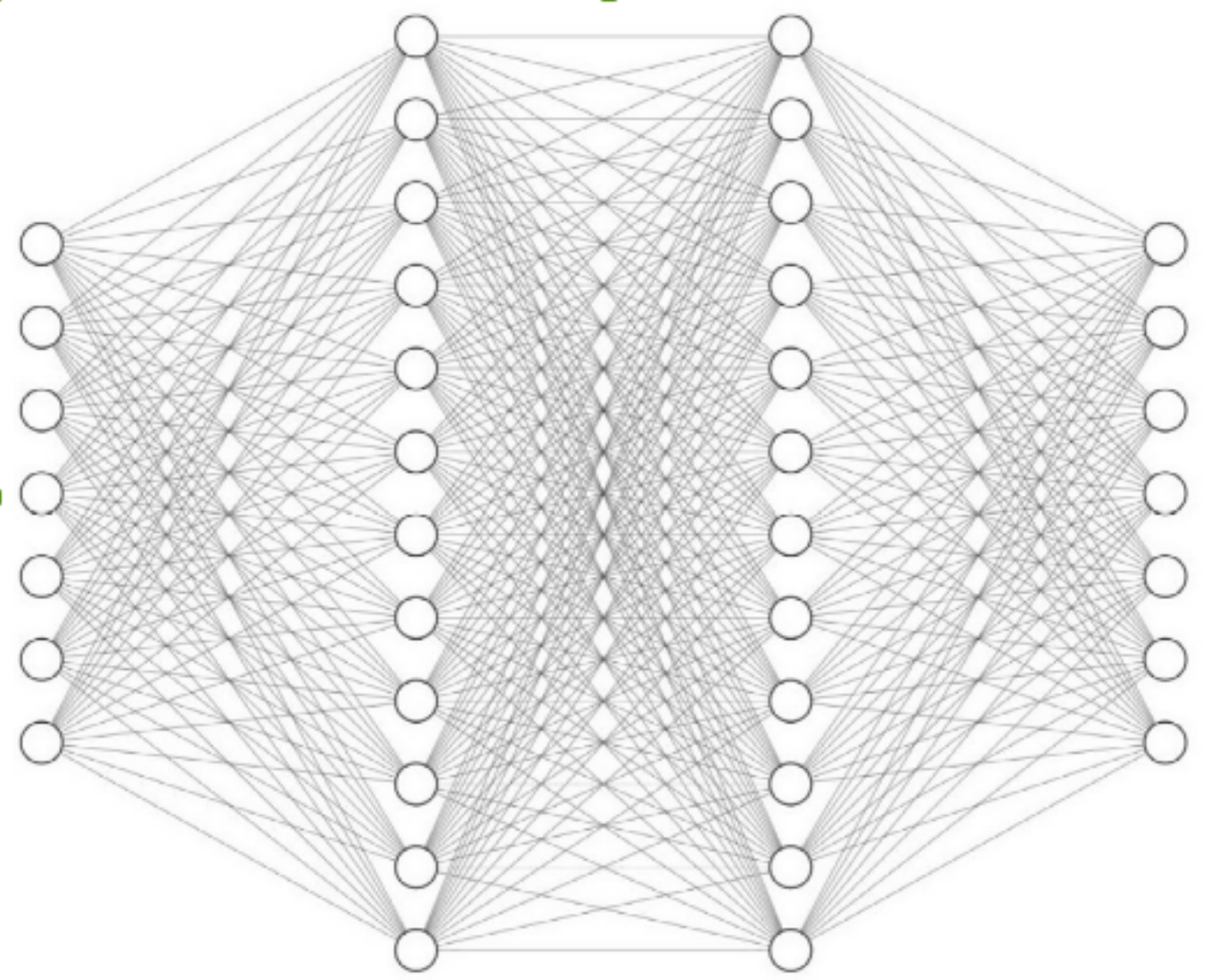}
  \centering
  \caption{Demonstration of a Feedforward Neural Network}
\end{figure}

The forward network depicted in Figure 12. shows a simplified version of the actual network used on this project, therefore, a more general description of the architecture is explained. As mentioned earlier, the dimensions of the boards to be trained were decided as 7x7 to maintain simplicity towards the neural network analysis while allowing room for exploration of more complex boards in the case of obtaining initial desirable results. Thus, the initial layer of the FNN consists of 49 neurons (7x7) each depicting a digit between 0-9. It is important to note that the digits were normalized as (-0.5 , 0.5) as neural networks show to perform better within this intervals. Following the input layers are 2 hidden layers which after multiple tests, the best and faster results obtained were for each hidden layer to include 1024 neurons with a ReLU activation function. Lastly, the output layer also contains 49 neurons (7x7), however, since a Softmax layer was applied to this problem, it must be noted that each of the 7x7 cells can take a digit from 0-9, therefore, the output cell consisted of 49*10 neurons. The loss function used was the classic "CrossEntropy" along with the classic "Stochastic Gradient Descent" optimization algorithm.    

\begin{figure}
    \centering
  \includegraphics[width=0.4\linewidth]{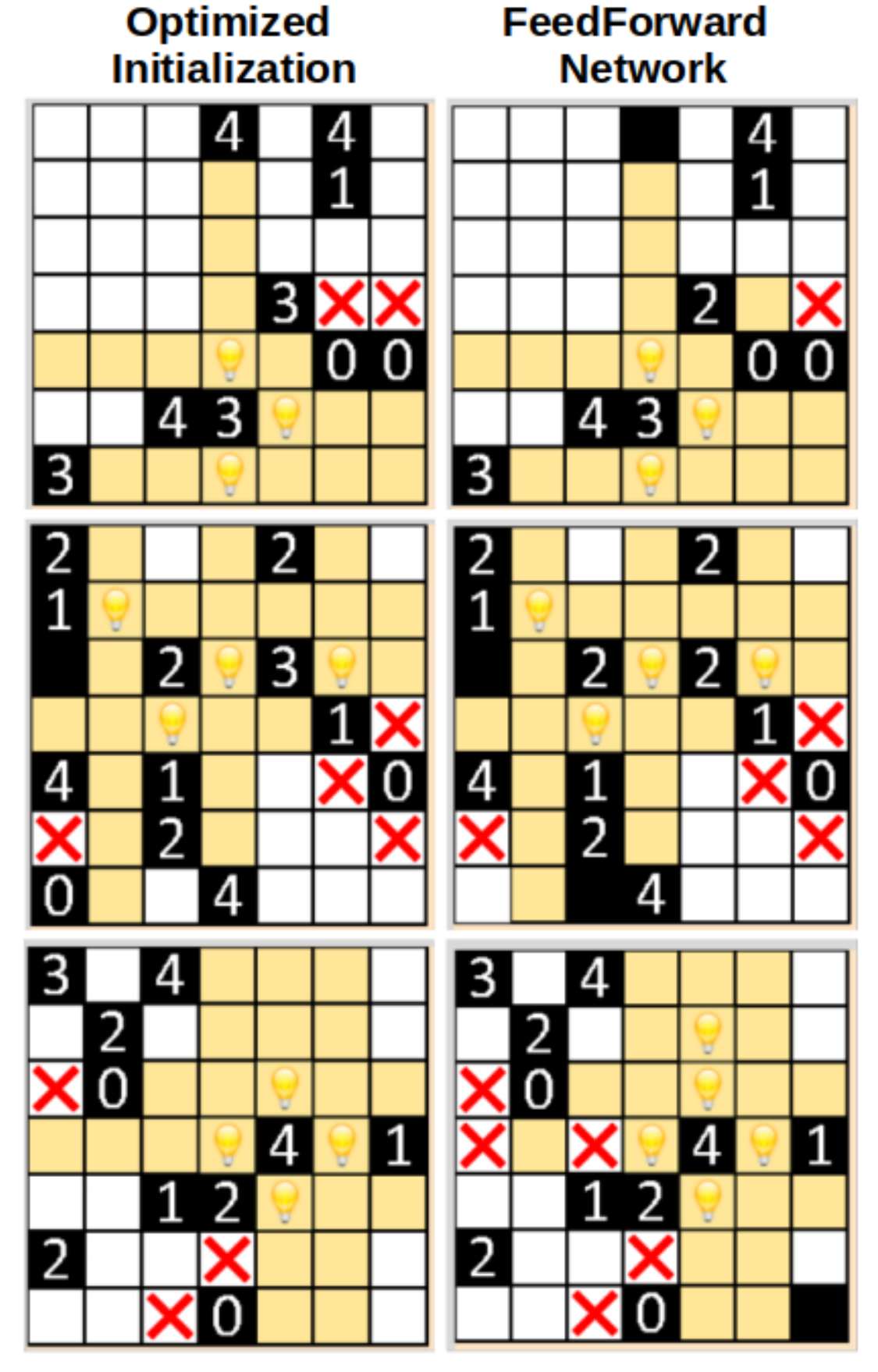}
  \centering
  \caption{Demonstration of the results from the FNN to predict initialized board}
\end{figure}

From Figure 13. it can be seen that the FNN does a poor job at predicting the initialized boards. Comparing the Vanilla Hill-Climb initialization on the left column, to the FNN initialization on the right column, some of the clear errors include varying of the initial board's black cells configuration or initial state, misplacement of light bulbs intersecting each other, and misplacement of crosses randomly throughout the board. These errors show to be crucial as they change entirely the configuration of the actual initial board. In addition to this, the FNN showed a major time set back for training taking approximately 1 hour and 30 minutes. Therefore, a new approach was necessary not only to test a fully solved board by the Hill-Climb algorithm once it was completed, but a new and faster neural network architecture or method.       

\subsection{\normalsize{Convolutional Neural Network (CNN) coupled to Vanilla Hill-Climb}}

Following the FNN tests on the initialization step, the full Vanilla Hill-Climb algorithm was completed shortly after. Therefore, a new dataset had to be developed where given an initial randomly generated board of dimensions 7x7 as the input to the network, the fully solved board could be used as the labeled set. The length of the dataset was kept as 5000, which took approximately 3-4 days to fully collect and upload to the Hopper supercomputer. 

\begin{figure}
    \centering
  \includegraphics[width=0.8\linewidth]{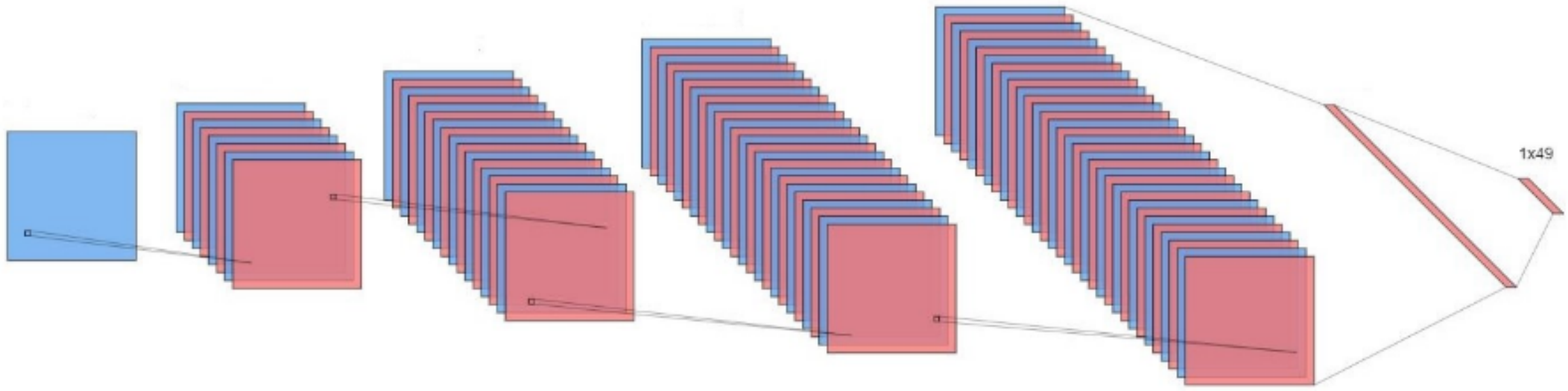}
  \centering
  \caption{Demonstration the CNN architecture developed}
\end{figure}

Once the dataset was gathered, the next step was to develop a new method of neural networks which could prove to be more effective and time efficient. Therefore, the idea was to implement a CNN approach, where instead of inputting a single large array of 49 neurons, a matrix of size 7x7 was used as the input resulting in a more intuitive network based to the board's configuration. After testing several CNN architectures, the best results were observed for the structure shown in Figure 14. 
The implemented CNN configuration consisted of the input as a 7x7 matrix with 1 single channel, followed by four "Same Padding" convolutions each with kernel size of 5 filters and a ReLU activation function. The channels were incremented from the input to the fourth convolution from 1, 8, 16, 32, 64, respectively. The output from the fourth convolution was then flatten using two DENSE layers of size 64*7*7 and 256 neurons, respectively. Lastly, unlike the FNN, after testing the CNN using a softmax output layer of 49*10 neurons the results were extremely incorrect predicting one single cell for the entirety of the board. This result could point out that FNN along with softmax layer and more advanced machine learning strategies might be an attractive approach to explore in future experiments rather than CNN with softmax. However, the output layer was reduced to only 49 neurons (7x7) similar to the input layer. The loss function used was changed to a regression problem with the Hubber Loss function (or SmoothL1Loss), and the optimization algorithm was the classic "Adam" optimizer. A key observation to make from this alteration of a one-hot-encoded logits prediction approach to a regression approach is that the outputs would not be integers from 0-9, rather they were floats (decimals). Therefore, during the training process the outputs were rounded to the nearest integer using Pytorch's built-in function $torch.round()$.

\begin{figure}
    \centering
  \includegraphics[width=1.0\linewidth]{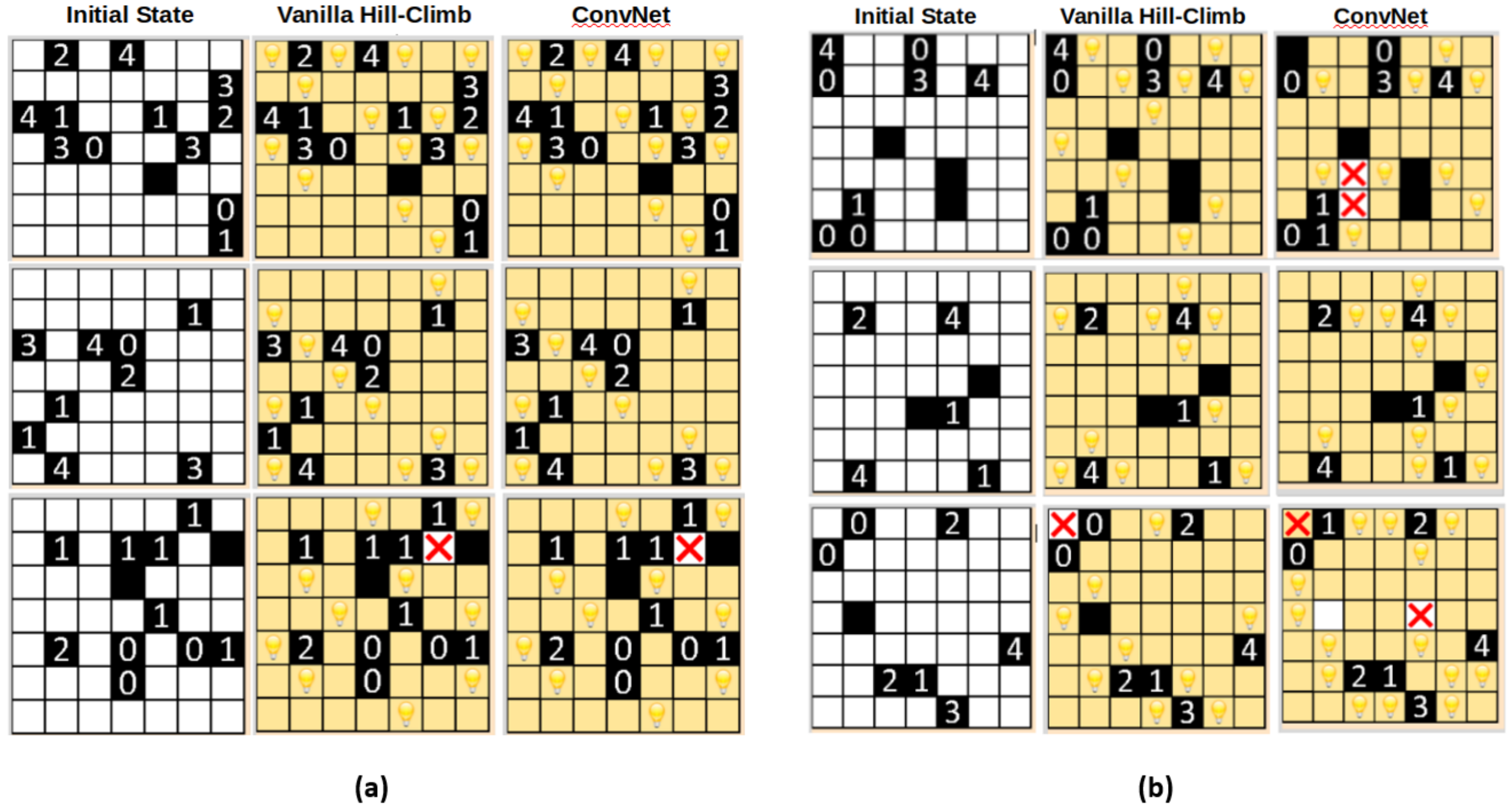} 
  \centering
  \caption{(a) Resultant initial predicted boards from the CNN method. (b) Testing the trained CNN for over-fitting on new random boards.}
\end{figure}

As shown in Figure 15a. the initial results from the CNN method showed identical predictions on the validation set. Although these results were encouraging to conclude that neural networks can solve the Light-Up puzzle, they can be misleading. One of the sources to think that these results cannot be used to make such a conclusion is the theory of over-fitting for neural networks \citep{overfitting}. Therefore, rather than conclude with these results, another test was performed by randomly generating new boards and testing the trained CNN network for over-fitting. Therefore, as seen in in Figure 15b. a clear depiction of over-fitting error is present where the new boards are not solved according to the rules of the Light-Up puzzle. Some of these errors are intersecting light bulbs, misplacing extra crosses, and leaving empty white cells. However, it can also be observed an improvement between the FNN results and the CNN results. The simple CNN is able to capture a stronger relationship by not altering the original board's configuration or initial state as much, as well as illuminating more cells which points out to the idea that it does slightly better at recognizing that white cells must end up being illuminated cells. As a result of these findings, a final attempt was necessary once the Simulated Annealing algorithm was developed. 

The idea that the Simulated Annealing would provide less chance for over-fitting is inspired by the fact that usually Simulated Annealing avoids the problem to get stuck in the local optima. However, it is important to emphasize once more on the fact that the Light-Up puzzle has such a large number of available moves, leading to a new global optima at each iteration. This observation is crucial in order to interpret the results obtained from the neural networks. In other words, a simple change from a data set of board solutions obtained using the Vanilla Hill-Climb algorithm, to a data set of solutions obtained using the Simulated Annealing algorithm might actually lead to the exact same solution when implementing CNNs. However, these conclusions were worth exploring, which lead our team to the following lasts approach using the Simulated Annealing.     

\subsection{\normalsize{Regularized CNN coupled to Simulated Annealing}}

The final analysis of CNN involved testing the Simulated Annealing results along with the inclusion of the dropout regularization technique to the CNN. The first step to this final approach was to obtain a data set for solved boards using the Simulated Annealing. This time, gathering the data took approximately 5-6 days due to the complexity of the algorithm as well as a larger data set generation of 10,000 random boards.

The architecture of the CNN was kept the same as the configuration used for the Vanilla Hill-Climb, however, 4 regularization dropout layers were added between each convolution, and one last dropout layer was added following the first DENSE flatten layer. The idea behind dropout regularization follows the theory that for every training iteration a number of neurons get randomly eliminated according to a probability defined by the user \citep{dropout}. This random elimination of neurons between each convolution and DENSE layer, in theory, should reduce the over-fitting from the model by eliminating dependence from the trained weights at each iteration. The probabilities used from the first convolutional layer to the first DENSE flatten layer were 0.1, 0.2, 0.25, 0.25, and 0.3, respectively.

\begin{figure}
\centering
  \includegraphics[width=1.0\linewidth]{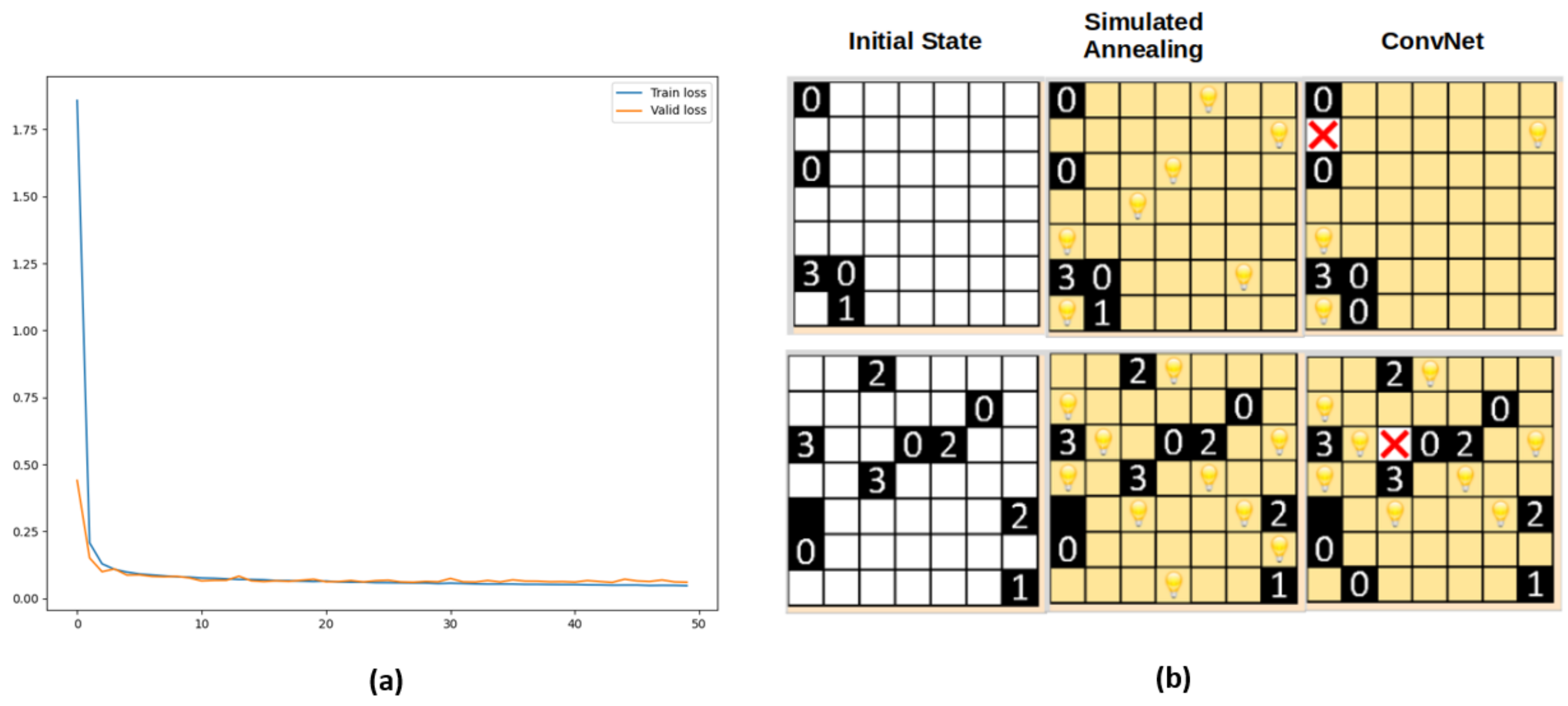}
  \caption{(a) Training Loss and Validation Loss as a function of training iterations.(b) Regularized CNN results on Simulated Annealing Data Set.}
  \centering
\end{figure}

As shown in Figure 16a., a collection of the training set loss and validation set loss was gathered through out the training process in order to visualize the overfitting results, if any. From these results we can see that the error on the training set converges to approximately 0.1 at around epoch No. 10. Similarly, the validation loss converges very close to the training error at around epoch No. 10. This points out to the fact that the number of epochs could be reduced to speed off training. 
However, when plotting the results from the regularized CNN on the simulated annealing validation data, it can be seen from Figure 16b., that the prediction is still not perfect. Although we have eliminated or reduced the over-fitting in the data, the errors are still visible in the prediction. Some of the errors we can observe are missing light bulbs, and one missing black cell. Nevertheless, taking away that the results are not perfect, there is a notable improvement when using the regularization technique on the CNN along with the Simulated Annealing approach. The new CNN is able to capture that each white cell must be filled with an illumination or a cross, as well as some of the predicted positions for light bulbs are exact to the simulated annealing approach.

\subsection{\normalsize{Conclusion of Neural Networks on the Light-Up Puzzle}}

In conclusion, the use of a regression problem approach using CNNs rather than a one-hot-encoded logits array using the softmax function proved to be a better approach for predicting the puzzles' solutions. However, an area to explore in future experiments is to develop more complex FNNs using the softmax approach on the Simulated Annealing results. Another hope for future experiments is to use more complex CNNs algorithms which include constraints. For instance, as mentioned earlier the "Sudoku Puzzle" uses a RNN approach by sequentially iterating through the entire board (cell by cell) in ascending order and predicting the number to input using a softmax layer of dimensions $board_{size}*9$, in combination with a constraint function recording the already filled cells. While this approach cannot simply be transferred to the Light-Up puzzle, there are some aspects which can be used as a starting point of improvement strategy. One idea can be to include a constraint function for each of the filled cells; this would reduce the number of digits needed to be predicted and ultimately eliminating the errors seen for FNN by varying the original initial state configuration. Additionally, instead of sequentially iterating through each cell in ascending order, a more intricate algorithm could be developed using some other rules for iterating through cells not in ascending order.  
This project developed new AI algorithms that not only proved to be more intuitive for solving the Light-Up puzzle, but showed their use in gathering data sets to be used in neural networks experiments. A future work involves the development of higher-complexity neural networks including constraint functions that model the rules of the Light-Up puzzle.

\section*{Supplementary Data}

The supplementary data used for the neural networks can be found on the GitHub repository: https://github.com/rperera12/AKARI-LightUp-GameSolver-with-DeepNeuralNetworks-and-HillClimb-or-SimulatedAnnealing.

\section*{Acknowledgments}

Authors are grateful for the support of the Auburn University Easley Cluster for assistance with this work.

\bibliographystyle{unsrtnat}
\bibliography{references}  %%% Uncomment this line and comment out the ``thebibliography'' section below to use the external .bib file (using bibtex) .

\begin{thebibliography}{9}
\providecommand{\natexlab}[1]{#1}
\providecommand{\url}[1]{\texttt{#1}}
\expandafter\ifx\csname urlstyle\endcsname\relax
  \providecommand{\doi}[1]{doi: #1}\else
  \providecommand{\doi}{doi: \begingroup \urlstyle{rm}\Url}\fi

\bibitem[McPhail(2005)]{first}
Brandon McPhail.
\newblock Light up is np-complete.
\newblock 01 2005.

\bibitem[Das and Chakrabarti(2005)]{third}
Arnab Das and Bikas Chakrabarti.
\newblock Quantum annealing and related optimization methods, lecture note in
  physics, vol. 679.
\newblock \emph{Quantum Annealing and Related Optimization Methods, Edited by
  A. Das and B.K. Chakrabarti. 2005 XIV, 378 p. 124 illus. Also available
  online. ISBN 3-540-27987-3. Berlin: Springer, 2005.}, 01 2005.

\bibitem[Weinberger(1990)]{fourth}
Edward Weinberger.
\newblock Correlated and uncorrelated fitness landscapes and how to tell the
  difference.
\newblock \emph{Biological Cybernetics}, 63:\penalty0 325--, 09 1990.
\newblock \doi{10.1007/BF00202749}.

\bibitem[Gunawardena(2011)]{t-test}
Nalika Gunawardena.
\newblock Choosing the correct statistical test in research.
\newblock \emph{Sri Lanka Journal of Child Health}, 40, 12 2011.
\newblock \doi{10.4038/sljch.v40i4.3838}.

\bibitem[Sevgen et~al.(2017)Sevgen, Arslan, and Samli]{sudoku}
Selcuk Sevgen, Emel Arslan, and Ruya Samli.
\newblock Solving sudoku puzzle with numbers recognized by using artificial
  neural networks.
\newblock \emph{IU-JEEE}, 17:\penalty0 3205--3211, 03 2017.

\bibitem[Salcedo-Sanz et~al.(2009)Salcedo-Sanz, Carro-Calvo, Ortiz-García,
  Pérez-Bellido, and Portilla-Figueras]{evolution}
Sancho Salcedo-Sanz, Leopoldo Carro-Calvo, Emilio Ortiz-García, Ángel
  Pérez-Bellido, and Antonio Portilla-Figueras.
\newblock A nested two-steps evolutionary algorithm for the light-up puzzle.
\newblock \emph{ICGA Journal}, 32:\penalty0 131--139, 09 2009.
\newblock \doi{10.3233/ICG-2009-32302}.

\bibitem[Palm et~al.(2017)Palm, Paquet, and Winther]{RNN}
Rasmus~Berg Palm, Ulrich Paquet, and Ole Winther.
\newblock Recurrent relational networks for complex relational reasoning.
\newblock \emph{CoRR}, abs/1711.08028, 2017.
\newblock URL \url{http://arxiv.org/abs/1711.08028}.

\bibitem[Owen(1994)]{overfitting}
A.~Owen.
\newblock Overfitting in neural networks.
\newblock 09 1994.

\bibitem[Corradi and Xvhito(2002)]{dropout}
Valcntina Corradi and Halbert Xvhito.
\newblock Regularized neural networks:.
\newblock 05 2002.

\end{thebibliography}





% \end{thebibliography}

\end{document}